\documentclass{article}

\PassOptionsToPackage{numbers, sort&compress}{natbib}
\usepackage{arxiv}

\usepackage[utf8]{inputenc} 
\usepackage[T1]{fontenc}    
\usepackage{hyperref}       
\usepackage{url}            
\usepackage{booktabs}       
\usepackage{amsfonts}       
\usepackage{nicefrac}       
\usepackage{microtype}      
\usepackage{lipsum}		
\usepackage{graphicx}
\usepackage{natbib}
\usepackage{doi}

\title{Local learning through propagation delays \\ in spiking neural networks}

\author{J\o{}rgen Jensen Farner$^{1}$, Ola Huse Ramstad$^{1,2}$, Stefano Nichele$^{1,3*}$, Kristine Heiney$^{1,4}$\thanks{Co-senior authors, email: \texttt{stefano.nichele@hiof.no} and \texttt{kristine.heiney@oslomet.no}} \\
	$^{1}$Department of Computer Science, Oslo Metropolitan University\\
	$^{2}$Department of Neuromedicine and Movement Science, Norwegian University of Science and Technology\\
	$^{3}$Department of Computer Science and Communication, \O{}stfold University College\\
	$^{4}$Department of Computer Science, Norwegian University of Science and Technology}
	
\renewcommand{\shorttitle}{\textit{arXiv} Template}

\hypersetup{
pdftitle={arXiv preprint: Beyond weight plasticity},
pdfsubject={q-bio.NC, q-bio.QM},
pdfauthor={J\o{}rgen Jensen Farner, Ola Huse Ramstad, Stefano Nichele, Kristine Heiney},
pdfkeywords={delay plasticity, local learning, spiking neural networks, Izhikevich neuron, generalized learning},
}

\begin{document}
\maketitle

\begin{abstract}
	We propose a novel local learning rule for spiking neural networks in which spike propagation times undergo activity-dependent plasticity.
  Our plasticity rule aligns pre-synaptic spike times to produce a stronger and more rapid response.
  Inputs are encoded by latency coding and outputs decoded by matching similar patterns of output spiking activity.
  We demonstrate the use of this method in a three-layer feedfoward network with inputs from a database of handwritten digits.
  Networks consistently improve their classification accuracy after training, and training with this method also allowed networks to generalize to an input class unseen during training.
  Our proposed method takes advantage of the ability of spiking neurons to support many different time-locked sequences of spikes, each of which can be activated by different input activations.
  The proof-of-concept shown here demonstrates the great potential for local delay learning to expand the memory capacity and generalizability of spiking neural networks.
\end{abstract}


\section{Introduction}

The brain has a great capacity for learning and memory, and the mechanisms that allow it to reliably and flexibly store information can provide new foundational mechanisms for learning in artificial networks.
Perhaps the most widely discussed mechanism associated with learning is Hebbian plasticity \citep{hebb-organization-of-behavior-1949,Markram2011}. 
This theory on neural learning states that when one neuron causes repeated excitation of another, the efficiency with which the first cell excites the second is increased.

The basic idea underlying Hebbian mechanisms is the brain's ability to change: local activity changes how neurons in a network communicate with each other, in turn affecting the overall behavior. 
In Hebbian plasticity, these changes are to the strength of connections between neurons. 
However, experimental observations \citep{Bucher2011, Grossman1979, Hatt1976, Luscher1994} have demonstrated that local activity can affect not only the \textit{strength} of connections but also the \textit{speed} with which action potentials travel between neurons. 
This alteration in transmission delays is likely an inherent part of how the brain learns and stores memories, as encoding information in time-locked sequences expands the computational capacity of a network \citep{Izhikevich2006}.

Local plasticity rules, such as spike-timing-dependent plasticity (STDP) \citep{Markram1997}, that change synaptic weights in an activity-dependent manner are of great interest in the context of unsupervised deep learning in deep spiking neural networks (SNNs) \citep{Tavanaei2019}. 
But why should plasticity in SNNs be confined to synaptic weights, when we are aware of a much richer repertoire of plastic changes that occur in the brain \citep{Gittis2006plasticity,Zhang2003plasticity,Hansel2001plasticity}?
Delay plasticity in neural networks has been explored, but the majority of studies have used supervised methods \citep{Schrawen2004, Wang2019, Taherkhani2015, Johnston2006}, with one noteworthy study using an unsupervised method to train only the readout layer of a reservoir \citep{Paugam-MoisyHelene2008}.

Here, we present our novel STDP analogue for local delay learning \citep{JorgenThesis}.
In this proposed learning rule, the timing of pre- and post-synaptic spikes influences the \textit{delay} of the connection rather than its weight, causing any subsequent spike transmission between a pair of neurons to occur at a different speed.
The main mechanism of our method is to better align all pre-synaptic spikes causally related to a post-synaptic spike, with the purpose of producing a faster and stronger response in the post-synaptic neuron.
We apply our developed delay learning method to the classification of handwritten digits \citep{LeCun2005mnist} in a simple proof-of-concept and demonstrate that training delays in a feedforward SNN is an effective method for information processing and classification.
Our networks consistently outperformed their untrained counterparts and were able to generalize their training to a digit class unseen during training.

\section{Delay learning in spiking neural networks}
\label{sec:delaylearning}

This section presents the novel activity-dependent delay plasticity method developed in this study and the encoding and decoding approaches of latency coding (LC) and polychronous group pattern (PGP) clustering used in our delay learning framework\footnote{Code available upon request.}.
The goal of our proposed learning method is to consolidate the network activity associated with similar inputs that constitute a distinct input class, so that the network will produce similar patterns of activity to be read out.
With this aim in mind, the delays of pre-synaptic neurons that together produce activity in a post-synaptic neuron are adjusted to better align the arrival of their spikes at the post-synaptic neuron.
Our framework was developed using Izhikevich regular spiking (RS) neurons.

\begin{figure}
    \centering
    \includegraphics[width=0.7\linewidth]{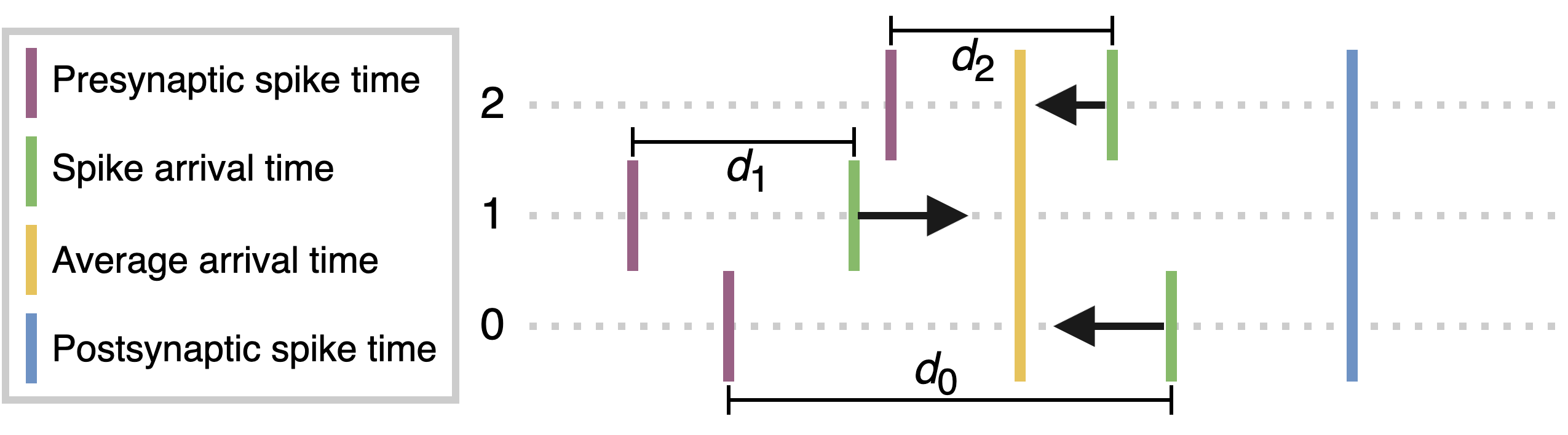}
    \caption{Schematic overview of the delay learning mechanism. Purple vertical lines indicate presynaptic spike initiation times, green lines indicate presynaptic spike arrival times according to their delays $d_i$, and the blue line indicates the post-synaptic spike time. The learning mechanism works by pushing pre-synaptic spikes that arrive before the post-synaptic spike towards their average arrival time, indicated by the yellow line.}
    \label{fig:spikealignment}
\end{figure}

Analogous to how STDP potentiates connections between causally related neurons to enhance the post-synaptic response, our delay plasticity mechanism increases the post-synaptic response by better aligning causally related pre-synaptic spikes.
This alignment process is illustrated in Fig.~\ref{fig:spikealignment} for the case of four pre-synaptic neurons connected to one post-synaptic neuron.
As shown in this figure, the pre-synaptic spikes (purple lines) that arrive (green lines) before the post-synaptic spike (blue line) are pushed towards their average arrival time (yellow line).
The delay $d_{i,j}$ between pre-synaptic neuron $i$ and post-synaptic neuron $j$ is changed according to the following equation:
\begin{equation}
    \label{eq:delayeq}
    \Delta d_{i,j} = 
		    -3 \, \mathrm{tanh}\left(\frac{t_i+d_{i,j}-\bar{t}_{\mathrm{pre}}}{3}\right), 
		    \;\;\;\; 0 \leq \Delta t_{\mathrm{lag}} < 10~\mathrm{ms},
\end{equation}
where $t_i$ is the spike time of neuron $i$, $\bar{t}_{\mathrm{pre}}$ is the average pre-synaptic arrival time across all neurons with spikes arriving within 10~ms before the post-synaptic spike, and $\Delta t_{\mathrm{lag}} = t_{j} - t_{i} + d_{i,j}$ is the time lag between when the pre-synaptic spike arrives at the post-synaptic neuron and when the post-synaptic neuron fires.
The time window of $10~\mathrm{ms}$ was selected because this is the window in which a pre-synaptic spike elicit a post-synaptic response.

\begin{figure}
    \centering
    \includegraphics[width=0.65\linewidth]{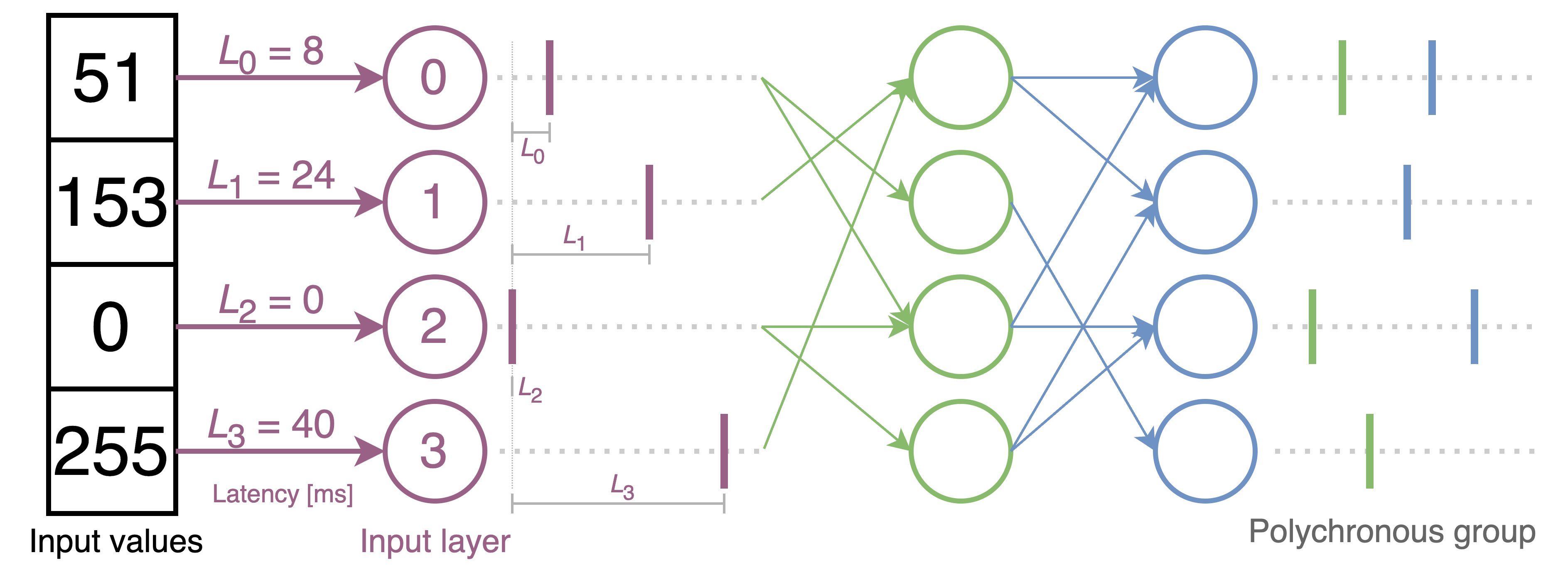}
    \caption{Illustration of the encoding and decoding methods. Left: Input values are encoded as spike latencies. Right: PGPs are defined as sets of sequential activity triggered by inputs, and they are clustered in a hierarchical manner by checking the ratio of matching spikes with other PGPs.}
    \label{fig:encodingdecoding}
\end{figure}

The encoding and decoding approaches are illustrated in Fig.\ \ref{fig:encodingdecoding}.
In LC, inputs are encoded in the relative spike timing of the input neurons.
That is, input channels with a value of $0$ will fire first, followed by other channels in order of increasing input value.
Through experimentation, we determined that rescaling the dynamic range to relative latencies of $[0,40~\mathrm{ms}]$ produced good results.
Our decoding approach of PGP clustering is based on the concept of polychronization, introduced by Izhikevich as the occurrence of ``reproducible time-locked but not synchronous firing patterns'' \citep{Izhikevich2006}.
A polychronous group is an ensemble of neurons that can produce multiple such time-locked PGPs depending on how they are activated.
Because inputs from the same class do not activate precisely the same input neurons, we also introduced a method of assigning distinct output PGPs to the same class.
In this PGP clustering method, we iteratively merge PGPs into clusters based on how closely the order of spikes matches the mean of all PGPs already in that cluster; the threshold for matching was set to 80\% and 90\% of the mean total spike count between the two PGPs being compared.

\section{Proof-of-concept: Classification of handwritten digits} 
\label{sec:classification}

To demonstrate the utility of our proposed delay learning method, we applied it to the classification of handwritten digits \citep{LeCun2005mnist}.
This dataset consists of images of $28 \times 28$ pixels; we scaled these images down to a size of $10 \times 10$ and assigned an input neuron to each pixel.
The details of our experimental setup are given in Table \ref{tab:architecture}.
We used feedforward networks with three layers, including the input layer, and fixed homogeneous connection weights.

\begin{table}
  \caption{Network architecture and experimental parameters}
  \label{tab:architecture}
  \centering
  \begin{tabular}{cccccccc}
    \toprule
    Layer  &  Number  &  Connection   &    &  Digits  &  Train   &  Test  & PGP match\\
    size  &  of layers  &  probability  &  Weight  &  (unseen)  &  instances  &  instances & threshold \\
    \midrule
    $100$  &  3  &   0.1  &  $6$  &  $0,1,(2)$  &  $20$  &  $25$ & $80$\%, $90$\% \\
    \bottomrule
  \end{tabular}
\end{table}

In each iteration of the experiment, a feedforward network was generated with connectivity between layers according to the connection probability and connections assigned random initial delays in the range of $(0,40~\mathrm{ms})$ (integer values with uniform probability).
We then provided inputs from the selected digit classes to this untrained network with local plasticity switched off to give a performance baseline for random delays.
In the training phase, different inputs of the same digit classes were fed into the network with local delay plasticity switched on.
Following training, we again switched off local plasticity and provided the same set of inputs as given in the baseline test phase to assess the performance of the trained network.
One digit class was selected as an ``unseen'' class, i.e., a class presented during testing but not training, to evaluate the network's ability to generalize.

\begin{figure}
    \centering
    \includegraphics[width=0.9\linewidth]{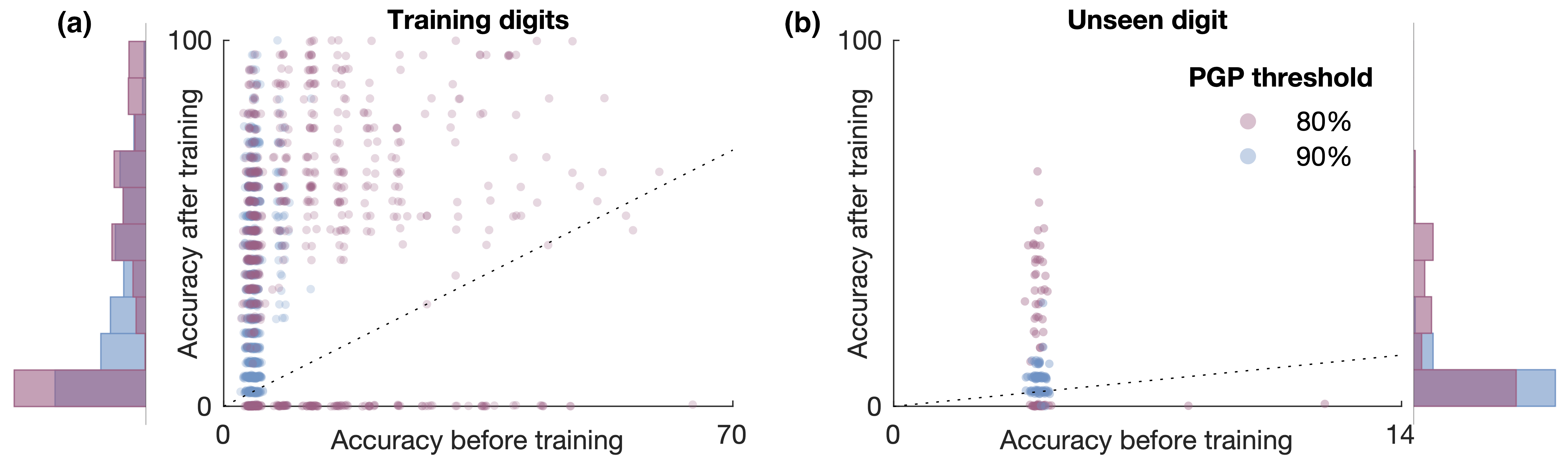}
    \caption{Results of classifying handwritten images of two digits before and after training using delay learning, for (a) 500 and (b) 100 networks initialized and tested with the parameters listed in Table \ref{tab:architecture}. Accuracy of classifying (a) two training digit classes (0,1) and (b) one unseen digit class (2). Results are plotted with jitter for the sake of visualization. Histograms show the number of networks with each given accuracy. Accuracy of 0 indicates non-separable classes.}
    \label{fig:accuracy}
\end{figure}

Fig.\ \ref{fig:accuracy} shows the accuracy before and after training,
calculated as the ratio of the count of the most common PGP class to the total presented inputs.
In nearly all cases where the network could separate the digit classes, the trained network performed better than the corresponding untrained network; however, some networks were unable to separate the classes (2.4\% and 45\% of networks for PGP thresholds $\theta=90$\% and $80$\%, respectively; see Fig.\ \ref{fig:accuracy}(a)).
Networks were also able to generalize their learning to a digit class unseen during training (Fig.\ \ref{fig:accuracy}(b)). 
Here, the accuracy remained low for the more stringent $\theta=90$\% but reached up to 64\% for $\theta=80$\% (mean accuracy 32\% in 38 networks able to separate the unseen class).
Flexibility with the PGP threshold can thus allow networks to generalize its training to unseen classes while maintaining good performance on trained classes.

\begin{figure}
    \centering
    \includegraphics[width=\linewidth]{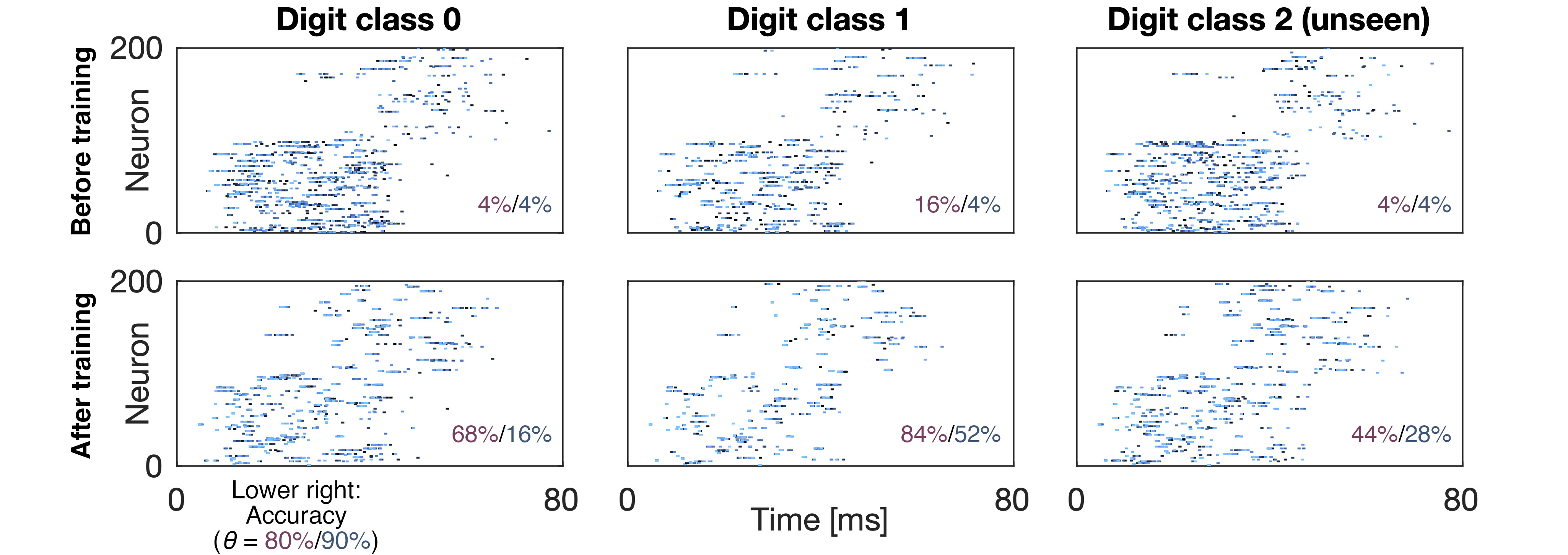}
    \caption{Raster plots of activity in layers 2 and 3 (neurons 1--100 and 101--200, respectively) before and after training. Digit classes 0 and 1 were used for training, and 2 is an unseen third class presented only during testing. Colors represent 25 different inputs from each class. Accuracies at PGP thresholds of 80\% and 90\% are reported in the lower right corner of each plot.}
    \label{fig:rasters}
\end{figure}

Examples of the activity in the output layers before and after training are shown in Fig.\ \ref{fig:rasters}.
This demonstrates the way the delay learning pushes the network to produce recognizably similar patterns (PGPs) when presented with inputs from the same class, as evidenced by the greater overlap of activity patterns after training.
Prior to training, the network activity is less structured overall and sparse in the final layer (neurons 101--200), whereas after training, the final layer is more active, and consistent spiking patterns can be observed across many inputs from the same class.

\section{Discussion}

Neural networks with carefully designed spike time delays can support many time-locked patterns of activity, expanding the coding capacity when compared with traditional rate models \citep{Izhikevich2006}.
Delay learning enables such polychronization in populations of spiking neurons, and our results show that we can take advantage of this richness of activity to train networks that can generalize their training to new inputs.
Our results demonstrate that feed-forward SNNs trained with our proposed local delay plasticity rule produce similar activity patterns in their output layers that can be well classified with a strict PGP matching threshold of 90\%.
Furthermore, lowering the threshold to 80\% yielded some networks able to generalize their training to novel inputs unseen during the training period.

Our proof-of-concept shows the great potential for this local delay learning method; even with only a short training period of 20 digit presentations, PGPs emerge in the network activity that allow for improved classification accuracy.
However, there remains much room for improvement.
In the cases where the network performs poorly, it is largely due to non-separability of input classes, and this is frequently accompanied by a fairly high accuracy prior to training (see Fig.\ \ref{fig:accuracy}(a) with threshold 80\%).
In these cases, the networks are likely being over-trained and producing a homogeneous PGP that reprepsents multiple groups.
An appropriate stopping point for training must be designed to avoid this pitfall.
In future work, networks can also be designed with heterogeneous weights and neuron types beyond the RS neuron.
In these cases, it may also be beneficial to apply the use of optimization techniques, such as evolutionary algorithms, to design effective network layouts.

\section*{Acknowledgements}

This work was partially funded by the SOCRATES project (Research Council of Norway, IKTPLUSS grant agreement 270961) and the DeepCA project (Research Council of Norway, Young Research Talent grant agreement 286558).

\bibliographystyle{unsrtnat}
\bibliography{references}  

\begin{thebibliography}{19}
\providecommand{\natexlab}[1]{#1}
\providecommand{\url}[1]{\texttt{#1}}
\expandafter\ifx\csname urlstyle\endcsname\relax
  \providecommand{\doi}[1]{doi: #1}\else
  \providecommand{\doi}{doi: \begingroup \urlstyle{rm}\Url}\fi

\bibitem[Hebb(1949)]{hebb-organization-of-behavior-1949}
Donald~O Hebb.
\newblock \emph{The organization of behavior: {A} neuropsychological theory}.
\newblock Wiley, New York, June 1949.
\newblock ISBN 0-8058-4300-0.

\bibitem[Markram et~al.(2011)Markram, Gerstner, and Sjöström]{Markram2011}
Henry Markram, Wulfram Gerstner, and Per~Jesper Sjöström.
\newblock A history of spike-timing-dependent plasticity.
\newblock \emph{Frontiers in synaptic neuroscience}, 3:\penalty0 4--4, 2011.
\newblock ISSN 1663-3563.

\bibitem[Bucher and Goaillard(2011)]{Bucher2011}
Dirk Bucher and Jean-Marc Goaillard.
\newblock Beyond faithful conduction: short-term dynamics, neuromodulation, and
  long-term regulation of spike propagation in the axon.
\newblock \emph{Progress in neurobiology}, 94\penalty0 (4):\penalty0 307--346,
  2011.
\newblock ISSN 0301-0082.

\bibitem[Grossman et~al.(1979)Grossman, Parnas, and Spira]{Grossman1979}
Yoram Grossman, Itzchak Parnas, and Micha~E Spira.
\newblock Differential conduction block in branches of a bifurcating axon.
\newblock \emph{The Journal of physiology}, 295\penalty0 (1):\penalty0
  283--305, 1979.
\newblock ISSN 0022-3751.

\bibitem[Hatt and Smith(1976)]{Hatt1976}
Hans Hatt and Dean~O Smith.
\newblock Synaptic depression related to presynaptic axon conduction block.
\newblock \emph{The Journal of physiology}, 259\penalty0 (2):\penalty0
  367--393, 1976.
\newblock ISSN 0022-3751.

\bibitem[L\"{u}scher et~al.(1994)L\"{u}scher, Streit, Quadroni, and
  L\"{u}scher]{Luscher1994}
Christian L\"{u}scher, J\"{u}rg Streit, Reto Quadroni, and Hans-Rudolf
  L\"{u}scher.
\newblock Action potential propagation through embryonic dorsal root ganglion
  cells in culture. i. influence of the cell morphology on propagation
  properties.
\newblock \emph{Journal of Neurophysiology}, 72\penalty0 (2):\penalty0
  622--633, 1994.
\newblock \doi{10.1152/jn.1994.72.2.622}.

\bibitem[Izhikevich(2006)]{Izhikevich2006}
Eugene~M. Izhikevich.
\newblock Polychronization: Computation with spikes.
\newblock \emph{Neural Comput}, 18\penalty0 (2):\penalty0 245--282, 2006.
\newblock ISSN 1530-888X,0899-7667.
\newblock \doi{10.1162/089976606775093882}.

\bibitem[Markram et~al.(1997)Markram, L\"{u}bke, Frotscher, and
  Sakmann]{Markram1997}
Henry Markram, Joachim L\"{u}bke, Michael Frotscher, and Bert Sakmann.
\newblock Regulation of synaptic efficacy by coincidence of postsynaptic aps
  and epsps.
\newblock \emph{Science}, 275\penalty0 (5297):\penalty0 213--215, 1997.
\newblock \doi{10.1126/science.275.5297.213}.

\bibitem[Tavanaei et~al.(2019)Tavanaei, Ghodrati, Kheradpisheh, Masquelier, and
  Maida]{Tavanaei2019}
Amirhossein Tavanaei, Masoud Ghodrati, Saeed~Reza Kheradpisheh, Timothée
  Masquelier, and Anthony Maida.
\newblock Deep learning in spiking neural networks.
\newblock \emph{Neural Networks}, 111:\penalty0 47--63, 2019.
\newblock ISSN 0893-6080.
\newblock \doi{10.1016/j.neunet.2018.12.002}.

\bibitem[Gittis and {du Lac}(2006)]{Gittis2006plasticity}
Aryn~H. Gittis and Sascha {du Lac}.
\newblock Intrinsic and synaptic plasticity in the vestibular system.
\newblock \emph{Current Opinion in Neurobiology}, 16\penalty0 (4):\penalty0
  385--390, 2006.
\newblock ISSN 0959-4388.
\newblock \doi{10.1016/j.conb.2006.06.012}.
\newblock Sensory systems.

\bibitem[Zhang and Linden(2003)]{Zhang2003plasticity}
Wei Zhang and David~J. Linden.
\newblock The other side of the engram: experience-driven changes in neuronal
  intrinsic excitability.
\newblock \emph{Nature Reviews Neuroscience}, 4\penalty0 (11):\penalty0
  885--900, 2003.
\newblock \doi{10.1038/nrn1248}.

\bibitem[Hansel et~al.(2001)Hansel, Linden, and D'Angelo]{Hansel2001plasticity}
Christian Hansel, David~J. Linden, and Egidio D'Angelo.
\newblock Beyond parallel fiber ltd: the diversity of synaptic and non-synaptic
  plasticity in the cerebellum.
\newblock \emph{Nature Neuroscience}, 4\penalty0 (5):\penalty0 467--475, 2001.
\newblock \doi{10.1038/87419}.

\bibitem[Schrauwen and van Campenhout(2004)]{Schrawen2004}
Benjamin Schrauwen and Jan van Campenhout.
\newblock Extending spikeprop.
\newblock In \emph{2004 IEEE International Joint Conference on Neural Networks
  (IEEE Cat. No.04CH37541)}, volume~1, pages 471--475, 2004.
\newblock \doi{10.1109/IJCNN.2004.1379954}.

\bibitem[Wang et~al.(2019)Wang, Lin, and Dang]{Wang2019}
Xiangwen Wang, Xianghong Lin, and Xiaochao Dang.
\newblock A delay learning algorithm based on spike train kernels for spiking
  neurons.
\newblock \emph{Frontiers in neuroscience}, 13:\penalty0 252--252, 2019.
\newblock ISSN 1662-4548.

\bibitem[Taherkhani et~al.(2015)Taherkhani, Belatreche, Yuhua, and
  Maguire]{Taherkhani2015}
Aboozar Taherkhani, Ammar Belatreche, Li~Yuhua, and Liam~P Maguire.
\newblock Dl-resume: A delay learning-based remote supervised method for
  spiking neurons.
\newblock \emph{IEEE Trans Neural Netw Learn Syst}, 26\penalty0 (12):\penalty0
  3137--3149, 2015.
\newblock ISSN 2162-237X.
\newblock \doi{10.1109/TNNLS.2015.2404938}.

\bibitem[Johnston et~al.(2006)Johnston, Prasad, Maguire, and
  McGinnity]{Johnston2006}
Simon~P Johnston, Girijesh Prasad, Liam~P Maguire, and T~Martin McGinnity.
\newblock A hybrid learning algorithm fusing stdp with ga based explicit delay
  learning for spiking neurons.
\newblock In \emph{2006 3rd International IEEE Conference Intelligent Systems},
  pages 632--637. IEEE, 2006.
\newblock ISBN 9781424401956.

\bibitem[Paugam-Moisy et~al.(2008)Paugam-Moisy, Martinez, and
  Bengio]{Paugam-MoisyHelene2008}
Hélène Paugam-Moisy, Régis Martinez, and Samy Bengio.
\newblock Delay learning and polychronization for reservoir computing.
\newblock \emph{Neurocomputing (Amsterdam)}, 71\penalty0 (7):\penalty0
  1143--1158, 2008.
\newblock ISSN 0925-2312.

\bibitem[Jensen~Farner(2022)]{JorgenThesis}
J\o{}rgen Jensen~Farner.
\newblock Activity dependent delay learning in spiking neural networks.
\newblock Master's thesis, Oslo Metropolitan University, 2022.

\bibitem[LeCun and Cortes(2005)]{LeCun2005mnist}
Yann LeCun and Corinna Cortes.
\newblock The mnist database of handwritten digits.
\newblock 2005.

\end{thebibliography}






\end{document}